\documentclass{article}

\usepackage[final]{corl_2017}

\usepackage[utf8]{inputenc} 
\usepackage[T1]{fontenc}    
\usepackage{url}            
\usepackage{booktabs}       
\usepackage{amsfonts}       
\usepackage{nicefrac}       
\usepackage{microtype}      

\usepackage{amsfonts}       
\usepackage{nicefrac}       
\usepackage{microtype}      
\usepackage{amsmath}
\usepackage{multirow}
\usepackage{tikz}
\usepackage{pbox}
\usepackage{enumitem}
\usepackage{graphicx}
\usepackage{bbm}
\usepackage{float}
\usepackage{xfrac}
\usepackage{array}
\usepackage{pifont}
\usepackage{dblfloatfix}
\usepackage{caption}
\usepackage{bm}

\usepackage{times}
\usepackage{graphicx} 
\usepackage{subfigure} 
\graphicspath{{figs/}}
\usepackage{gensymb}

\usepackage[prependcaption,textsize=tiny]{todonotes}

\newcolumntype{L}[1]{>{\raggedright\let\newline\\\arraybackslash\hspace{0pt}}m{#1}}
\newcolumntype{C}[1]{>{\centering\let\newline\\\arraybackslash\hspace{0pt}}m{#1}}
\newcolumntype{R}[1]{>{\raggedleft\let\newline\\\arraybackslash\hspace{0pt}}m{#1}}

\DeclareCaptionLabelFormat{andtable}{#1~#2  \&  \tablename~\thetable}

\title{Learning End-to-end Multimodal Sensor Policies \\ \protect for Autonomous Navigation}

\usepackage{authblk}
 
\author[1]{Guan-Horng Liu}
\author[2]{Avinash Siravuru}
\author[1]{Sai Prabhakar}
\author[1]{ \protect\\ Manuela Veloso}
\author[1]{George Kantor}
\affil[ ]{\texttt{\string{guanhorl,asiravur,spandise\string}@andrew.cmu.edu}}
\affil[ ]{\texttt{mmv@cs.cmu.edu, kantor@ri.cmu.edu}}
\affil[1]{Robotics Institute, Carnegie Mellon University, USA}
\affil[2]{Department of Mechanical Engineering, Carnegie Mellon University, USA}

\begin{document}

\maketitle

\begin{abstract}
Multisensory polices are known to enhance both state estimation and target tracking. However, in the space of end-to-end sensorimotor control, this multi-sensor outlook has received limited attention. Moreover, systematic ways to make policies robust to partial sensor failure are not well explored. In this work, we propose a specific customization of Dropout, called \textit{Sensor Dropout}, to improve multisensory policy robustness and handle partial failure in the sensor-set. We also introduce an additional auxiliary loss on the policy network in order to reduce variance in the band of potential multi- and uni-sensory policies to reduce jerks during policy switching triggered by an abrupt sensor failure or deactivation/activation. Finally, through the visualization of gradients, we show that the learned policies are conditioned on the same latent states representation despite having diverse observations spaces - a hallmark of true sensor-fusion. Simulation results of the multisensory policy, as visualized in TORCS racing game, can be seen here:
\url{https://youtu.be/QAK2lcXjNZc}.
\end{abstract}

\section {Introduction}
One of the key challenges in building robust autonomous navigation systems is the development of a strong intelligence pipeline that is able to efficiently gather incoming sensor data and take suitable control actions with good repeatability and fault-tolerance. 
In the past, this was addressed in a modular fashion, where specialized algorithms were developed for each sub-system and integrated with  fine tuning. More recent trends show a revival of end-to-end approaches that learn complex mappings directly from the input to the output by leveraging large volume of task-specific data and the remarkable abstraction abilities afforded by deep neural networks. In autonomous navigation, these techniques have been used for learning visuomotor policies \cite{nvidiacar} from human driving data.
However, the traditional deep supervised learning-based driving requires a great deal of human annotation, and yet, may not be able to deal with the problem of accumulating errors during test time \cite{ross2011reduction}. On the other hand, deep reinforcement learning (DRL) offers a better formulation that allows policy improvement with feedback, and has achieved human-level performance on challenging game environments \cite{mnih2013playing, mnih2015human}.

In this work, we present an end-to-end controller that uses multi-sensor input to learn an autonomous navigation policy in a physics-based gaming environment called TORCS \cite{wymann2000torcs} (without needing any pretraining). To show the effectiveness of multisensory perception, we pick two popular continuous action DRL algorithms namely Normalized Advantage Function (NAF) \cite{CDQN} and Deep Deterministic Policy Gradient (DDPG) \cite{DBLP:journals/corr/LillicrapHPHETS15}, and augment them to accept multisensory input. We limit our objective to only achieving autonomous navigation without any obstacles or other cars. This problem is kept simpler to focus on analyzing the performance of the proposed multi-sensor configurations using extensive quantitative and qualitative testing. 
Sensor redundancy can be a bane if the policy relies heavily on all inputs and lead to significant performance drop even if a single sensor fails. In order to avoid this situation, we apply a customized stochastic regularization technique called \emph{Sensor Dropout} during training. Our approach reduces the policy over-dependence on a specific sensor subset, and guarantees minimal performance drop even in the face of any partial sensor failure. We further augment the standard DRL loss with an additional auxiliary loss to reduce variance in the trained policy and offer smoother performance during abrupt sensor loss or re-activation. 

Recently, promising experimental results were shown combining camera and lidar to build an end-to-end steering controller of a UGV navigation \cite{patelsensor}. Similarly, a multimodal DQN was built for a Kuka YouBot \cite{bohez2017sensor} by fusing information for homogeneous sensing modalities. However, the fusion stage in \cite{patelsensor} is limited to sensors that are spatially redundant with each other, and requires the feature embedding of each sensor to have the same dimensionality. On the other hand, \cite{bohez2017sensor} requires a two-stage training scheme which first approximates a $Q^*$ function and then refines the policy with DropPath \cite{bohez2017sensor} regularization. In addition to longer training time, this only if you assume DropPath during the second stage doesn't throw the policy outside of the initially optimized policy distribution. Any two stage policy with regularization in the second stage has to make this strong assumption. 

The proposed method can be best seen as a far more generalized version of the above two. Multi-sensor fusion can be performed on heterogeneous sensing modalities, any where in the network pipeline, and in shorter timescales. Moreover, the objective is not only improving sensor-fusion but also providing guaranteed operation feature even if a sensor subset fails (unique to this work). Through extensive empirical testing we show the following exciting results in this paper:
\begin{enumerate}
\item Multisensory DRL with Sensor Dropout (SD) reduces performance drop in a noisy environment from $\approx 50\%$ to just $10\%$, when compared to a baseline system.
\item A multisensory policy with SD guarantees functionality even in a face a sensor subset failure. This particular feature underscores the need for redundancy in a safety-critical application like autonomous navigation.
\end{enumerate}

\section{Related Work}
Multisensory DRL aims to leverage the availability of multiple, potentially imperfect, sensor inputs to improve learned policy. Most autonomous driving vehicles today are equipped with an array of sensors like GPS, Lidar, Camera, and Odometer, etc. However, some of these sensors, like GPS and odometers, are readily available but seldom included in deep supervised learning models \cite{nvidiacar}. Even in DRL, policies are predominantly single sensor-based, i.e., either low-dimensional physical states, or high-dimensional pixels. For autonomous driving where it is essential to achieve highest possible safety and accuracy targets, developing policies that operate with multiple inputs is better suited. In fact, multisensory perception was an integral part of autonomous navigation solutions and even played a critical role in their success \cite{multimodaltartan} before the advent of deep learning based approaches. Sensor fusion offers several advantages, namely robustness to individual sensor noise/failure, improving object classification and tracking \cite{cho2014multi}, etc. In this light, several recent works in DRL have tried to solve the complex robotics tasks such as human-robot-interaction \cite{qureshi2016robot}, manipulation \cite{levine2016end} and maze navigation \cite{mirowski2017a} with multisensory sensor inputs. \citeauthor{mirowski2017a} use similar using similar sensory data as in this work to navigate through a maze. However, the robot evolves with simpler dynamics and the depth information is only used to formulate an auxiliary loss and \emph{not} as an input to learn a navigation policy.

Multisensory deep learning, popularly called Multimodal deep learning, is an active area of research in other domains like audiovisual systems \cite{ngmultimodal}, text/speech and language models \cite{srivastava2012multimodal}, etc. However, Multi-modal learning is conspicuous by its absence in the modern end-to-end autonomous navigation literature. Another challenge in multimodal learning is the specific case of over-fitting where instead of learning the underlying latent target state representation using multiple diverse observations, the model instead learns a complex representation in the original space itself, defeating the purpose of using multi-sensor observations and making the process computationally burdensome. An illustrative example for this case is a car navigating when all sensors remain functional but fails to navigate at all even if one sensor fails or is partially corrupted. This kind of behavior is detrimental and suitable regularization measures should be set up during training to avoid it.

Stochastic regularization is an active area of research in deep learning made popular by the success of, \textit{Dropout} \cite{dropout}. Following this landmark paper, numerous extensions were proposed  to further generalize this idea 
(\cite{blockout,dropconnect,zoneout,dropall}).
In the similar vein, an interesting technique has been proposed for specialized regularization in the multimodal setting namely ModDrop \cite{moddrop}. ModDrop, however, requires pretraining with individual sensor inputs using separate loss functions. The method is originally designed for multimodal deep learning on a \textit{fixed} dataset. We argue that for DRL where the training dataset is generated during run-time, pretraining for each sensor policy may end up optimizing on \textit{different} input distributions. In comparison, \emph{Sensor Dropout} is designed to be applicable to the DRL setting. With SD, a network can be directly constructed in an end-to-end fashion and the sensor fusion layer can be added just like Dropout. The training time is much shorter and scales better with increasing number of sensors.

\section{Multimodal Deep Reinforcement Learning}
\textbf{Deep Reinforcement Learning (DRL) Brief Review: }
We consider a standard Reinforcement Learning (RL) setup, where an agent operates in an environment ${E}$. At each discrete time step $t$, the agent observes a state $s_t \in \mathcal{S}$, picks an action $a_t \in \mathcal{A}$, and receives a scalar reward $r(s_t, a_t) \in \mathbb{R}$ from the environment. The return $R_t = \sum^T_{i=t} \gamma^{(i-t)}r(s_i,a_i)$ is defined as total discounted future reward at time step $t$, with $\gamma$ being a discount factor $\in [0,1]$. The objective of the agent is to learn a policy that eventually maximizes the expected return.
The learned policy, $\pi$, can be formulated as either stochastic $\pi(a|s) = \mathbb{P}(a|s)$, or deterministic $a = \mu(s)$. The value function $V^{\pi}$ and action-value function $Q^{\pi}$ describe the expected return for each state and state-action pair upon following a policy $\pi$. 
Finally, an advantage function $A^{\pi}(s_t,a_t)$ is defined as the additional reward or advantage that the agent will have for executing some action $a_t$ at state $s_t$ and it is given by $A^{\pi}(s_t,a_t) = Q^\pi(s_t, a_t) - V^\pi(s_t)$. 

In high dimensional state/action space, these functions are usually approximated by a suitable parametrization. Accordingly, we define $\theta^Q$, $\theta^V$, $\theta^A$, $\theta^\pi$, and $\theta^\mu$ as the parameters for approximating $Q$, $V$, $A$, $\pi$, and $\mu$ functions, respectively. It was generally believed that using non-linear function approximators would lead to unstable learning in practice. Recently, Mnih et al. \cite{mnih2013playing} applied two novel modifications, namely \textit{replay buffer} and \textit{target network}, to stabilize the learning with deep nets. Later, several variants were introduced that exploited deep architectures and extended to learning tasks with continuous actions \cite{DBLP:journals/corr/LillicrapHPHETS15,A3C,CDQN}. 

To exhaustively analyze the effect of multi-sensor input and the new stochastic regularization technique, we pick two algorithms in this work namely DDPG and NAF. It is worth noting that the two algorithms are very different, with DDPG being an off-policy actor-critic method and NAF an off-policy value-based one. By augmenting these two algorithms, we highlight that any DRL algorithm, modified appropriately, can benefit from using multi-sensor inputs. Due to space constraint, we list the formulation of the two algorithms in Supplementary Material (Section \ref{sec:supply_a}).

\begin{figure}[t]
\begin{center}
\centerline{\includegraphics[width=\columnwidth,trim= 0 850 0 30, clip=true]{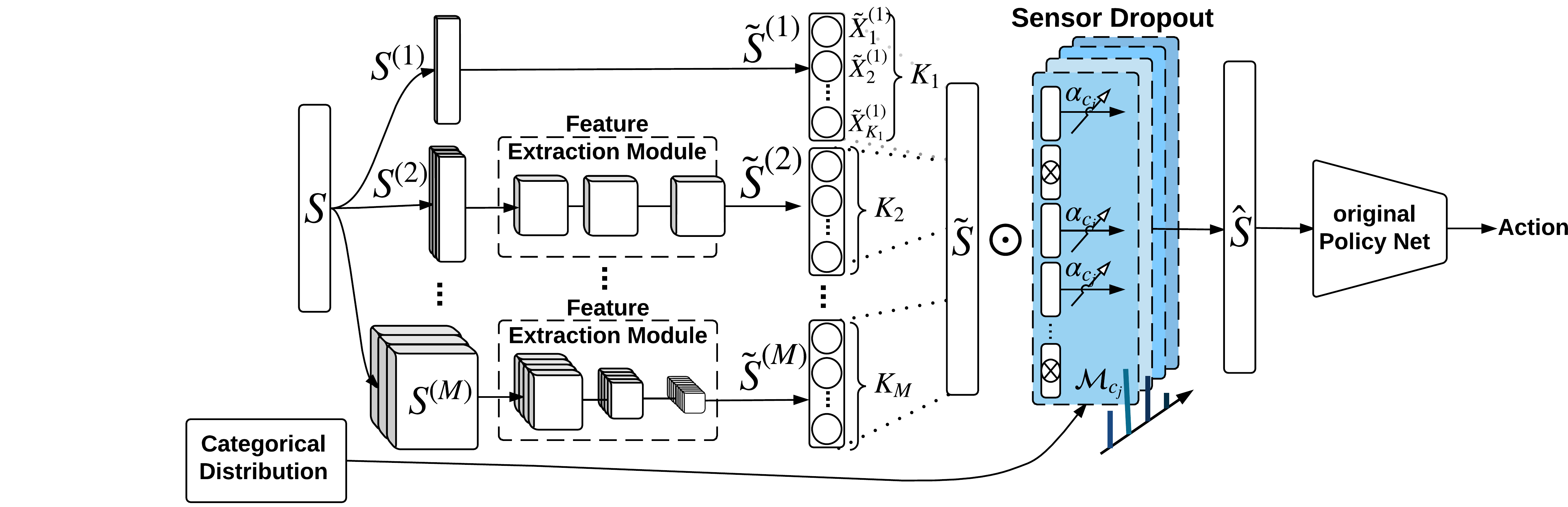}}
\caption{Illustration of multimodal sensor policy augmented with Sensor Dropout. The operation $\odot$ stands for element-wised multiplication. The dropping configuration of Sensor Dropout is sampled from a categorical distribution, which stands as an additional input to the network.
}
\label{fig:Multi-SD}
\end{center}
\vskip -0.3in
\end{figure}

\textbf{Multimodal (or) Multisensory Policy Architecture: }
We denote a set of observations composed from $M$ sensors as, $S = [S^{(1)}~S^{(2)}~..~S^{(M)}]^T$, where $S^{(i)}$ stands for observation from $i^{th}$ sensor. In the multimodal network, each sensory signal is pre-processed along an independent path. Each path has a feature extraction module that can be either pure identity function (modality $1$), or convolution-based layer (modality $2 \to M$). 
The modularized feature extraction stage naturally allows for independent extraction of salient information that is transferable (with some tuning if needed) to other applications
. 
The outputs of feature extraction modules are eventually flattened and concatenated to form the multimodal state. The schematic illustration of modularized multimodal policy is shown in Fig. \ref{fig:Multi-SD}.

\section{Augmenting MDRL}
In this section, we propose two methods to improve training of a multi-sensor policy. We first introduce a new stochastic regularization called Sensor Dropout, and explain its advantages over the standard Dropout for this problem. Later, we propose an additional unsupervised auxiliary loss function to reduce the policy variance. 

\textbf{Sensor Dropout (SD) for Robustness: } 
Sensor Dropout is a customization of Dropout \cite{dropout} that maintains dropping configurations on each sensor module instead of each neuron. Though both methods serve the purpose of regularization, SD is better-motivated for training multisensory policies. By randomly dropping the sensor block during training, the policy network is encouraged to exploit cross connections across different sensing streams. When applied to complex robotic system, SD has advantages of handling imperfect sensing conditions such as latency, noise and even partial sensor failure.
As shown in Fig.\ref{fig:Multi-SD}, consider the multimodal state $\tilde{S}$ 
, the dropping configuration is defined as a $M$-dimensional vector $\bm{c} = [\delta_{c}^{(1)}~\delta_{c}^{(2)}~..~\delta_{c}^{(M)}]^T$, where each element $\delta_{c}^{(i)} \in \{0,1\}$ represents the on/off indicator for the $i^{th}$ sensor modality. 
Each sensor modality is represented by a $K_i$-dimensional vector, denoted as $\tilde{S}^{(i)}= [\tilde{X}_1^{(i)}~\tilde{X}_2^{(i)}~..~\tilde{X}_{K_i}^{(i)}]^T$. The subscript $i$ indicates that each sensor may have different dimension.
We now detail the two main differences between original Dropout and SD along with their interpretations. 

Firstly, note that the dimension of the dropping vector $\bm{c}$ is much lower than the one in the standard Dropout ($\sum_{i=1}^M K_i$). As a consequence, the probability of the event where all sensors are dropped out (i.e. $\bm{{c_0}} = [0^{(1)}~0^{(2)}~..~0^{(M)}]^T$) is not negligible in SD. To explicitly remove $\bm{{c_0}}$, we slightly depart from \cite{dropout} in modeling the SD layer. Instead of modeling SD as random process where any sensor block $\tilde{S}^{(i)}$ is switched on/off with a \textit{fixed} probability $p$, we define the random variable as the dropping configuration $\bm{c}$ itself. Since there are $N = 2^M - 1$ possible states for $\bm{c}$, we accordingly sample from an $N$-state categorical distribution $\mathbb{P}$. 
We denote the probability of a dropping configuration $\bm{{c_j}}$ occurring with $p_j$, where the subscript $j$ ranges from $1$ to $N$. 
The corresponding pseudo-Bernoulli
\footnote{ We wish to point out that $p^{(i)}$ is pseudo-Bernoulli as we restrict our attention to cases where at least one sensor block is switched on at any given instant. This implies that switching-on of any sensor block $\tilde{S}^{(i)}$ is independent of the other but switching-off is not. So the distribution is no longer fully independent.}
distribution for switching on a sensor block $\tilde{S}^{(i)}$ can be calculated as $p^{(i)} = \sum_{j=1}^N\delta_{c_j}^{(i)} p_j$. 

\textbf{Remark:} Note that sampling from standard Bernoulli on sensor blocks with rejection of $c_0$ will have the same effect. However, the proposed categorical distribution aids in better bookkeeping and makes configurations easy to interpret.  It can also be adaptive to the current sensor reliability during run-time. 

Another difference from the standard Dropout is the rescaling process. Unlike the standard Dropout which preserves a \textit{fixed} scaling ratio after dropping neurons, the rescaling ratio in SD is formulated as a function of the dropping configuration and sensor dimensions. The intuition is to keep the weighted summations equivalent among different dropping configurations in order to activate the later hidden layers. The scaling ratio is calculated as $\alpha_{c_j} = \frac{\sum_{i=1}^M K_i }{\sum_{i=1}^M \delta_{c_j}^{(i)} K_i}.$ 

In summary, the output of SD for the $k^{th}$ feature in $i^{th}$ sensor block (i.e. $\tilde{S}^{(i)}$) given a dropping configuration $\bm{c_j}$ can be shown as $\hat{S}^{(i)}_{{c_j},k} = \mathcal{M}^{(i)}_{c_j} \tilde{X}_k^{(i)}$, 
where $\mathcal{M}^{(k)}_{c_j}=\alpha_{c_j} \delta_{c_j}^{(i)}$ is an augmented mask encapsulating both dropout and re-scaling. 




\textbf{Auxiliary Loss for Variance Reduction: } \label{sec:SD}
An alternative interpretation of the SD-augmented policy is that sub-policies induced by each sensor combination are jointly optimized during training. Denote the ultimate SD-augmented policy and sub-policy induced by each sensor combination as $\mu_{\bm{c}\sim \mathbb{P}}$ and $\mu_{c_j}$, respectively. The final output maintains a geometric mean over $N$ different actions. 

Though the expectation of the total policy gradients for each sub-policy is the same, SD provides no guarantees on the consistency of these actions. 
To encourage the policy network to extract salient features from each sensor that embed into a common latent state representation, we further add an auxiliary loss that penalizes the inconsistency among $\mu_{c_j}$. 
This additional penalty term provides an alternative gradient that reduces the variation of the ultimate policy, i.e. $Var \left[ \mu_{\bm{c}\sim \mathbb{P}} \right]$.
The mechanism is motivated from the recent successes \cite{mirowski2017a,DBLP:journals/corr/JaderbergMCSLSK16} that use the auxiliary tasks to improve both agent's performance and convergence rate. However, unlike most previous works that design the auxiliary tasks carefully from the ground truth environment, we formulate the \textit{target action} from the policy network itself.
Under the standard actor-critic architecture, the target action is defined as the output action of the sub-policy 
in target actor network $\tilde{\mu}_{\bm{c}\sim \mathbb{P}}$ that maximizes the target critic values $\tilde{Q}$. 
In other words, we use the currently best-trained sub-policy as a heuristic to guide other sub-policies during training.
\begin{align}
L_{aux} = \lambda \sum_{i=1}^N (\mu_{c_j}(s_i)-\tilde{\mu}_{c^{*}}(s_i))^2
\text{, \quad where  } c^{*} = \mathop{\mathrm{argmax}}_{c_j \sim \mathbb{P}} \sum_{i=1}^N \tilde{Q}(s_i,\tilde{\mu}_{c_j}(s_i))
\end{align}
Here, $\lambda$ is an additional hyperparameter that indicates the ratio between the two losses, and $N$ is the batch size for off-policy learning.

\section{Evaluation Results}
\subsection{Platform Setup} \label{sec:platform}
\textbf{TORCS Simulator}
The proposed approach is verified on TORCS \cite{wymann2000torcs}, a popular open-source car racing simulator that is capable of simulating physically realistic vehicle dynamics as well as multiple sensing modalities \cite{GymTORCS} to build sophisticated AI agents. 
In order to make the learning problem representative of the real-world setting, we use the following sensing modalities for our state description: (1) We define \emph{Sensor 1} as a hybrid state containing physical-based information such as odometry and simulated GPS signal. (2) \emph{Sensor 2} consists of $4$ consecutive laser scans (i.e., at time $t$, we input scans from times $t,~ t-1,~t-2~\&~t-3$). Finally, as \emph{Sensor 3}, we supply $4$ consecutive color images capturing the car's front-view. These three representations are used separately to develop our baseline uni-modal sensor policies. The multi-modal state, on the other hand, has access to all sensors at any given point. When Sensor Dropout (SD) is applied, the agent will randomly lose access to a strict subset of sensors. The categorical distribution is initialized with a uniform distribution among total $7$ possible combinations of sensor subset, and the best-learned policy is reported here. The action space is a continuous vector in $\mathbb{R}^2$, whose elements represent steering angle, and acceleration. Experiment details such as exploration strategy, network architectures of each model, and sensor dimensionality are shown in the Supplementary Material (Section \ref{sec:supply_b}).

\subsection{Results} \label{sec:results}


\begin{figure}[t]
\centering
\subfigure[NAF]{\label{fig:training_exp_naf}\includegraphics[width=0.4\columnwidth]{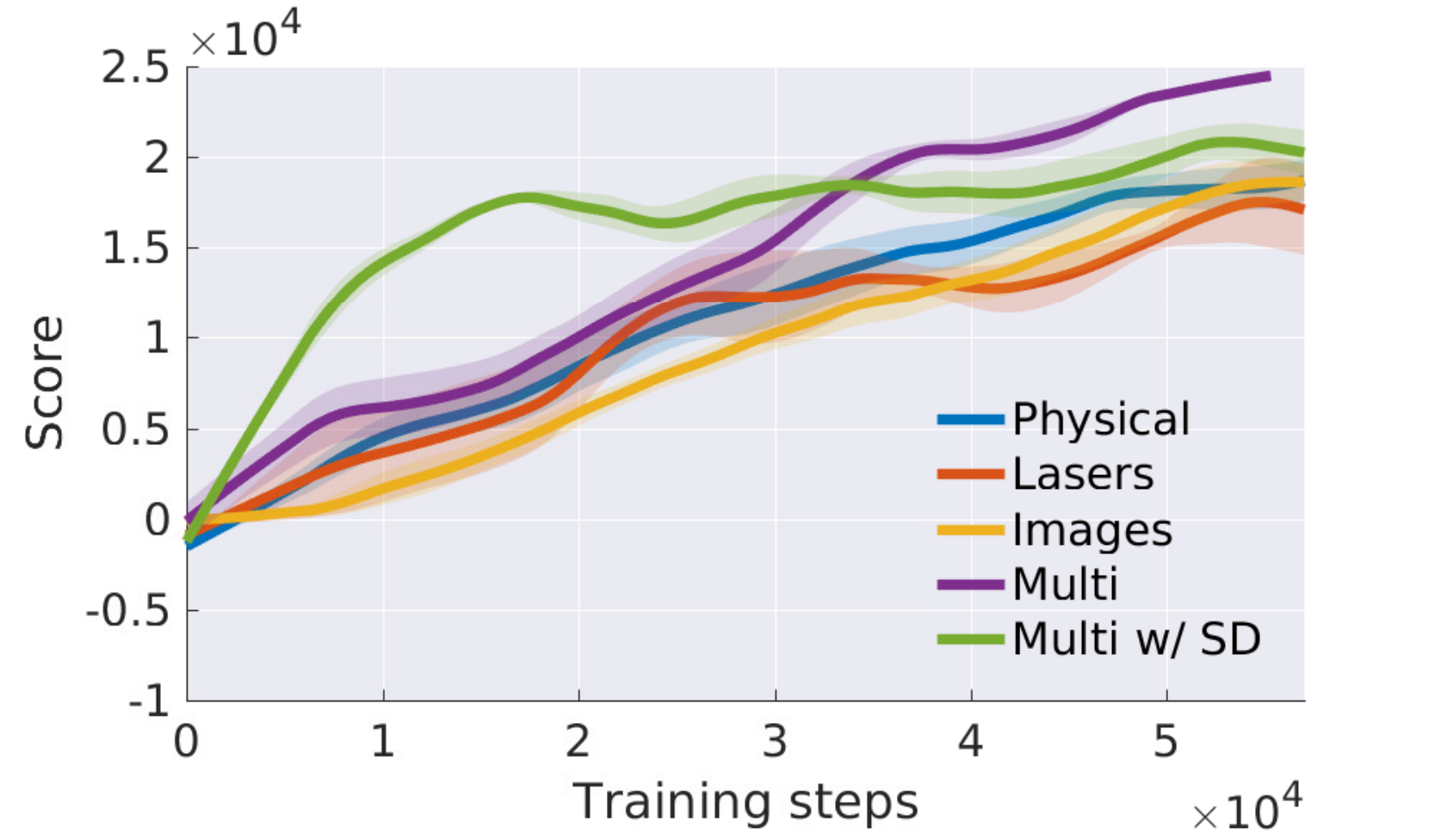}}
\subfigure[DDPG]{\label{fig:training_exp_ddpg}\includegraphics[width=0.4\columnwidth]{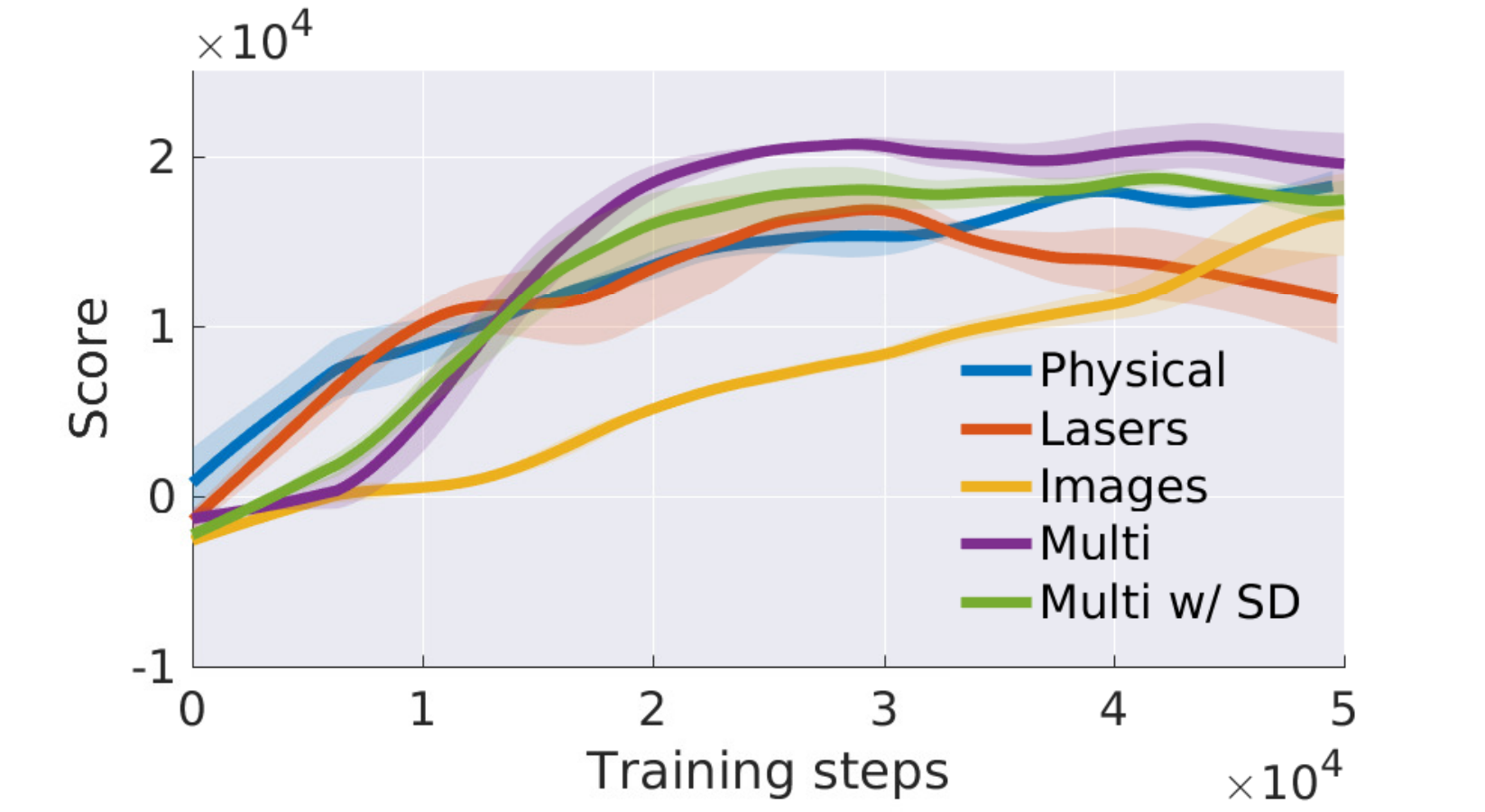}}
\caption{Training performance comparison of three baseline single sensor policies, and the proposed multi-modal policies, with and without Sensor Dropout.}
\label{fig:training_exp}
\vskip -0.25in
\end{figure}

\textbf{Training Summary:}
The training performances, for all the proposed models and their corresponding baselines, are shown in Fig. \ref{fig:training_exp}. 
For DDPG, using high-dimensional sensory input directly impacts convergence rate of the policy. Note that the \textit{Images} uni-policy (orange line) has a much larger dimensional state space compared with \textit{Multi} policies (purple and green lines).
Counter-intuitively, NAF performs a nearly linear improvement over training steps, and is relatively insensitive to the dimensionality of the state space. However, adding Sensor Dropout (SD) dramatically increases the convergence rate.
For both algorithms, the final performance for multimodal sensor policies trained with SD is slightly lower than training without SD, indicating that SD has a regularization effect similar to original Dropout.

\begin{table}[t]
	\vskip -0.1in
	\caption{Final Score of Trained Policy (\textit{unit}:$\times 10^4$) }
	\label{table:multiple-uni-baseline}
	\vskip 0.1in
    \centering
    \begin{small}
    \begin{sc}
    \begin{tabular}{c|cc|c}
    \toprule 
    \centering
    Policy & w/o Noise & w/ Noise & Performance Drop \\ \midrule \midrule
     Multi Uni-modal w/ Meta Controller & 1.51 $\pm$ 0.57 & 0.73 $\pm$ 0.40 & 51.7 \% \\
     Multimodal w/ SD & 2.54 $\pm$ 0.08 & 2.29 $\pm$ 0.60 & 9.8 \% \\ \toprule
    \end{tabular}
    \end{sc}
    \end{small}
\end{table}

\textbf{Comparison with Uni-modal Policies + Meta Controller:}
One of the intuitive baseline for the multi-sensor problem is to train each uni-modal sensor policy separately. Once individual policies are learned, we can train an additional meta-controller that select which policy to follow given the current state.
For this, we follow the setup in \cite{meta_policy} by training a meta controller that takes the processed states from each uni-modal policy, and outputs a $3 DOF$ softmax layer as the probability of choosing which sub-policy to perform.
Note that, we assume perfect sensing during the training. However, to test performance in a more realistic scenario, we simulate mildly imperfect sensing by adding Gaussian noise. Policy performance with and without noise are summarized in Table \ref{table:multiple-uni-baseline}. 
The performance of the baseline policy drops dramatically once noise is introduced, which implies that the uni-modal policy is prone to over-fitting without any regularization. In fact, the performance drop is sometimes severe in physical-based or laser-based policy. 
In comparison, the policy trained with SD reaches a higher score in both scenarios, and the drop when noise is introduced is almost negligible.
 


\begin{table}[t]
  \begin{minipage}[b]{0.45\linewidth}
    \centering
    \vskip -0.1in
    \includegraphics[width=\columnwidth]{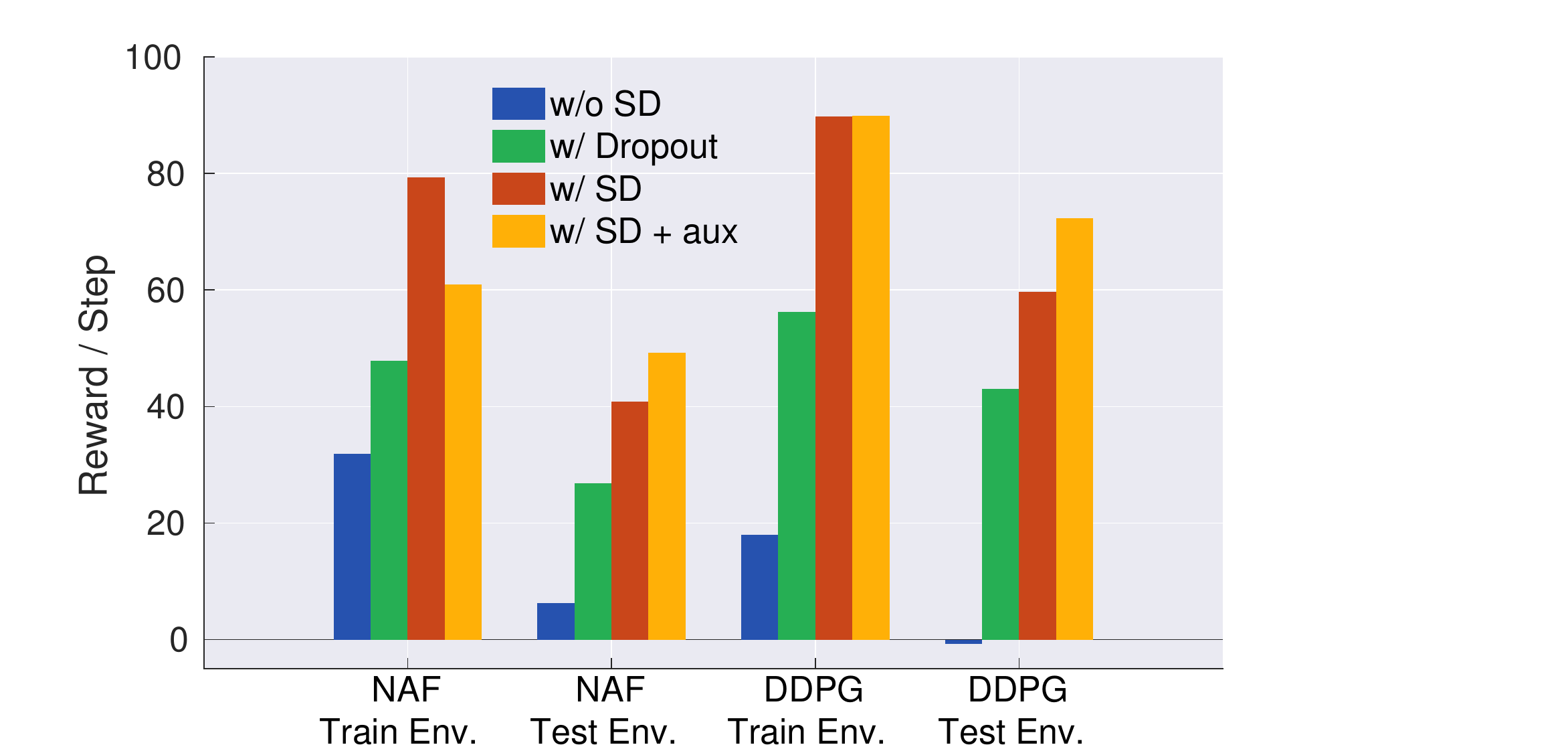}
    \vskip 0.1in
    \captionof{figure}{Policy performance when \\ facing random sensor failure.}
    \label{fig:random_sensor_failure}
  \end{minipage}
  \begin{minipage}[b]{0.45\linewidth}
  	\vskip -0.1in
    \caption{Results of the sensitivity metric.} 
	\label{table:policy-ratio}
    \vskip 0.1in
	\centering
    \begin{small}
    \begin{sc}
    \begin{tabular}{cccc}
    \toprule 
    \centering
     & & Training & Testing  \\ 
     & & Env. & Env. \\ \midrule \midrule
    \multirow{2}{*}{NAF}  & w/o SD & 1.651 & 1.722 \\
                          & w/ SD  & \textbf{1.284} & \textbf{1.086} \\ \midrule
    \multirow{2}{*}{DDPG} & w/o SD & 1.458 & 1.468 \\
                          & w/ SD  & \textbf{1.168} & \textbf{1.171} \\ \toprule
    \end{tabular}
    \end{sc}
    \end{small}
  \end{minipage}
  \vskip -0.1in
\end{table}

\textbf{Policy Robustness Analysis:}
In this part, we show that SD reduces the learned policy's acute dependence on a subset of sensors in a multimodal sensor setting. First, we consider a scenario when malfunctions of sensors have been detected by the system, and the agent must rely on the remaining sensors to make navigation decisions. To simulate this setting during testing, we randomly block out some sensor modules, and scale the rest using the same rescaling mechanism as proposed in Section \ref{sec:SD}. Fig. \ref{fig:random_sensor_failure} reports the averaging normalized reward of each model. A naive multimodal policy without any stochastic regularization (blue bar) performs poorly in the face of partial sensor failure and transfer tasks. Adding original Dropout makes the policy more generalized, yet the performance is not comparable with SD. Interestingly, by reducing the variance of the multimodal sensor policy with the auxiliary loss, policy tends to have a better generalization among other environments.

\textbf{Policy Sensitivity Analysis:}
To monitor the extent to which the learned policy depends on each sensor block, we measure the gradient of the policy output w.r.t a subset block $\tilde{S}^{(i)}$. The technique is motivated from the salient map analysis \cite{simonyan2013deep}, which has also been applied to DRL study recently \cite{WangFL15}. 
To better analyze the effects of SD, we report on a smaller subset by implementing SD layer to drop either (1) $(physical,~ laser)$ or (2) $vision$. Consequently, the \emph{sensitivity} metric is formulated as the relative sensitivity of the policy on two sensor subsets. If the ratio increases, the agent's dependence shifts toward the sensor block in the numerator and vice versa. Assuming the fusion-of-interest is between the above-mentioned two subsets, we show in Table \ref{table:policy-ratio} that, using SD, the metric gets closer to $1.0$, indicating nearly equal importance to both the sensing modalities. The \textit{sensitivity metric} is calculated as 
$\mathcal{T}_2^1 = \frac{1}{M}\sum_{i} \left( {\left| \nabla_{\tilde{S}^{(1)}_i} \mu (\tilde{S} | \theta^\mu )\Big|_{S_i} \right|} \right) \left( {\left| \nabla_{\tilde{S}^{(2)}_i} \mu (\tilde{S} | \theta^\mu )\Big|_{S_i} \right|} \right)^{-1} $.

\begin{figure}[t]
\vskip -0.1in
\centering
\includegraphics[width=0.8\columnwidth]{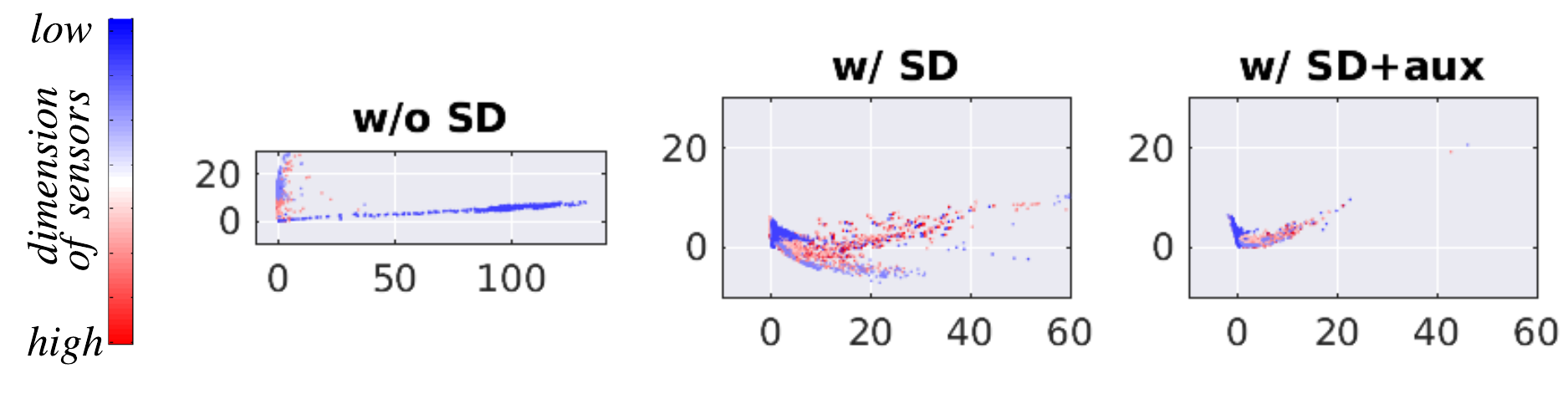}
\caption{Two-dimensional PCA embedding of the representations in the last hidden layer assigned by the policy networks. The blue dots correspond to the representations induced by the sub-policy that use high dimensional sensor (e.g. vision) as its input. On the other hand, the red dots represent the one with lower sensor stream such as odometry and range finder.}
\label{fig:aux_pca}
\vskip -0.2in
\end{figure} 

\begin{figure}[t]
\centering
\includegraphics[width=0.8\columnwidth]{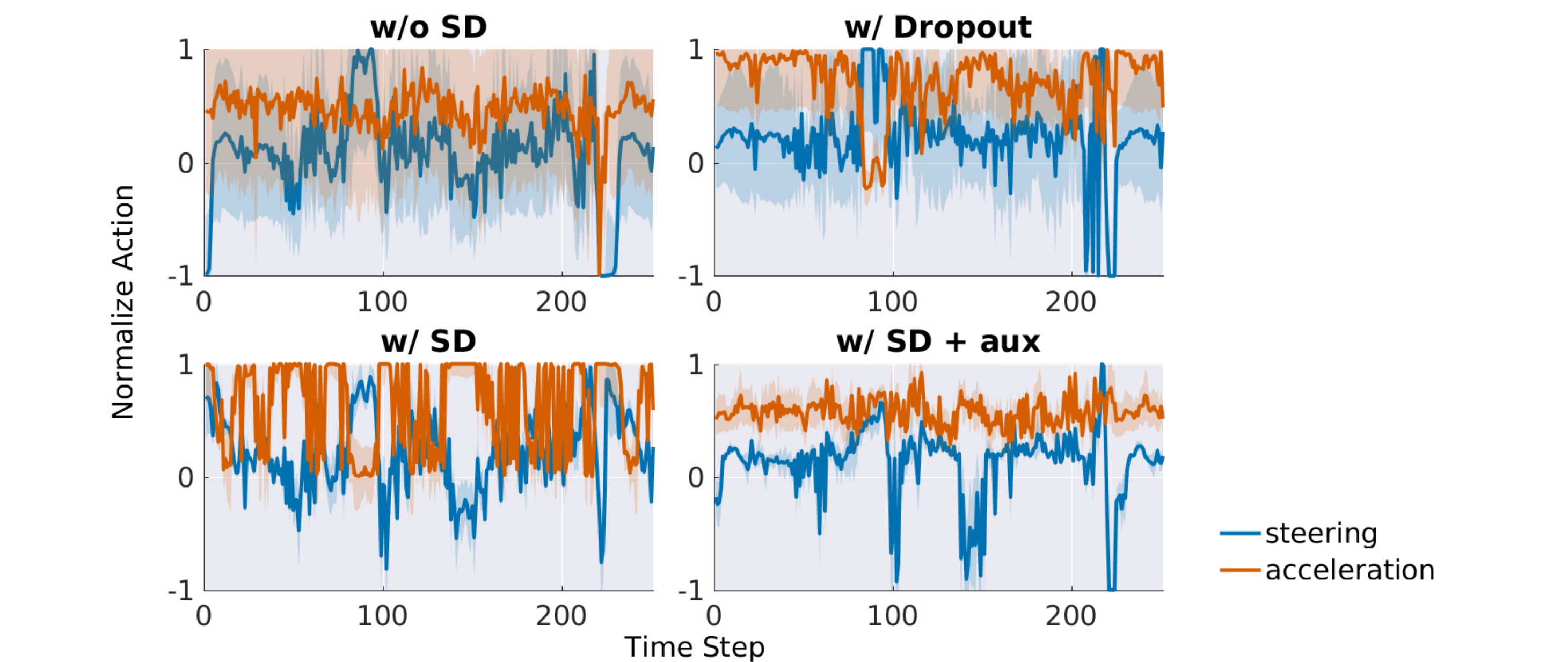}
\caption{The variance of all the actions induced by sub-policy under each multimodal sensor policy. \textit{Upper-left}: naive policy without any regularization. \textit{Upper-right}: with standard Dropout. \textit{Lower-left}: with Sensor Dropout. \textit{Lower-right}: with Sensor Dropout and auxiliary loss.}
\label{fig:action-profile}
\vskip -0.1in
\end{figure} 

\textbf{Effect of Auxiliary Loss:}
In this experiment, we verify how the auxiliary loss helps reshape the multimodal sensor policy and reduce the action variance. 
We extract the representations of the last hidden layer assigned by the policy network throughout a fixed episode. At every time step, the representation induced by each sensor combination is collected. 
Our intuition is that this latent space represents how the policy network interprets the incoming sensor stream for reaction. Based on this assumption, an ideal multimodal sensor policy should map different sensor streams to a similar distribution as long as the information provided by each combination is representative to lead to the same output action.

As shown in Fig. \ref{fig:aux_pca}, the naive multimodal sensor policy has a scattered distribution over the latent space, indicating that representative information from each sensor is treated very differently. In comparison, the policy trained with SD has a concentrated distribution, yet it is still distinguishable w.r.t. different sensors. Adding the auxiliary training loss encourages the true sensor fusion as the distribution becomes more integrated. During training, the policy is not only forced to explicitly make decisions under each sensor combination, but also penalized with the disagreements among multimodal sensor policies. In fact, as shown in Fig. \ref{fig:action-profile}, the concentration of the latent space directly affect the action variance induced by each sub-policy. 
We provide the actual covariances for each component and the actual action variance values in the Supplementary Material (Section \ref{sec:supply_c}).

\section{Discussion}

\begin{figure}[t]
\vskip -0.1in
\centering
\includegraphics[width=0.9\textwidth]{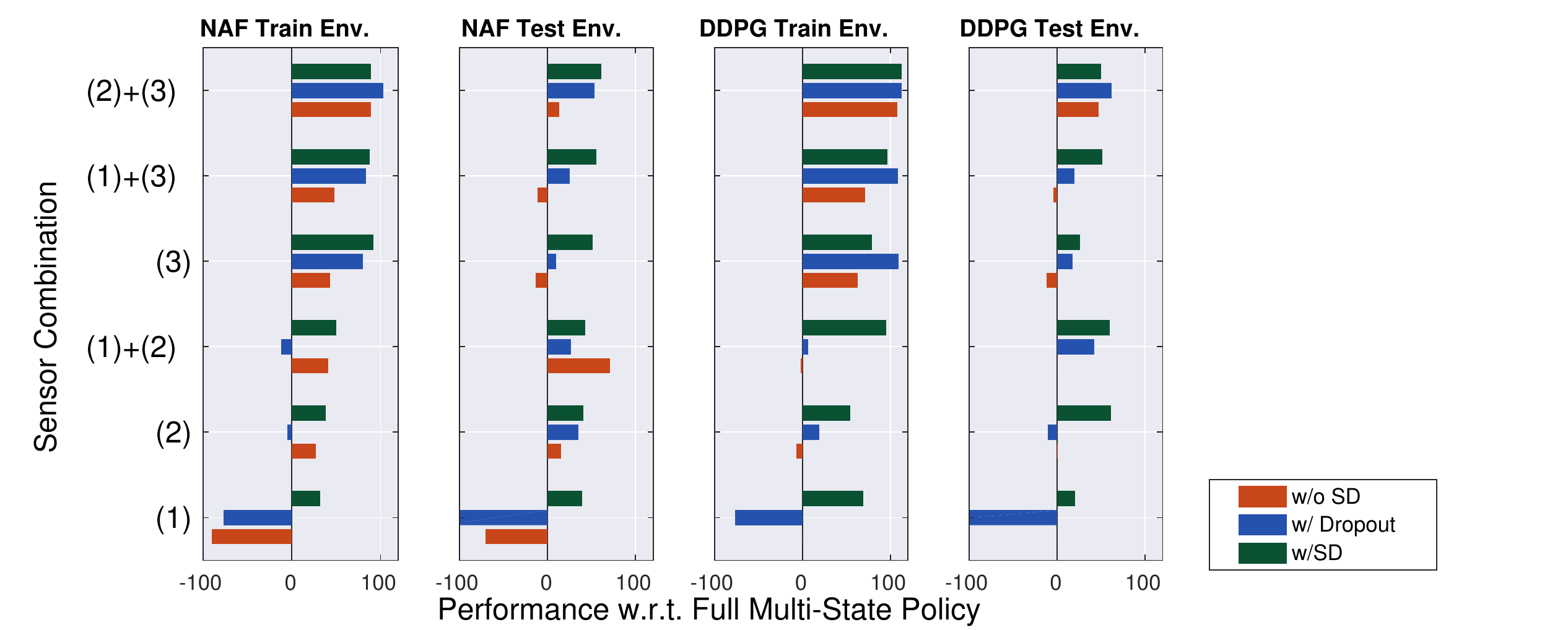}
\caption{The full analysis of the performance of the total $6$ sub-policies. The (1), (2), and (3) labels in y-axis represent physical state, laser, and image, respectively. The x-axis represent the remaining performance w.r.t. the SD policy with all sensor, i.e. (1)+(2)+(3).}
\label{fig:full-sd-policy}
\vskip -0.1in
\end{figure} 

\textbf{Full Sub-Policy Analysis: }
The performance of each sub-policy is summarized in Fig. \ref{fig:full-sd-policy}. 
As shown in the first and third column, the performances of the naive multimodal sensor policy (red) and the policy trained with standard Dropout (blue) drop dramatically as the policies lose access to image, which shares $87.9\%$ of the total multimodal state. Though Dropout increases the performance of the policy in the testing environment, the generalization is limited to using full multimodel state as input. On the other hand, SD generalizes the policy across \textit{sensor module}, making the sub-policies successfully transfer to the testing environment. It is worth mentioning that the policies trained with SD is capable to operate even when both laser and image sensor are blocked.
Interestingly, neither original Dropout or SD show apparent degradation in full policy induced by the regularization. 
We list more analysis as our future work.
 


\begin{figure}[t]
\vskip -0.1in
\centering
\subfigure[]{\label{fig:grad_exp}\includegraphics[width=0.48\columnwidth]{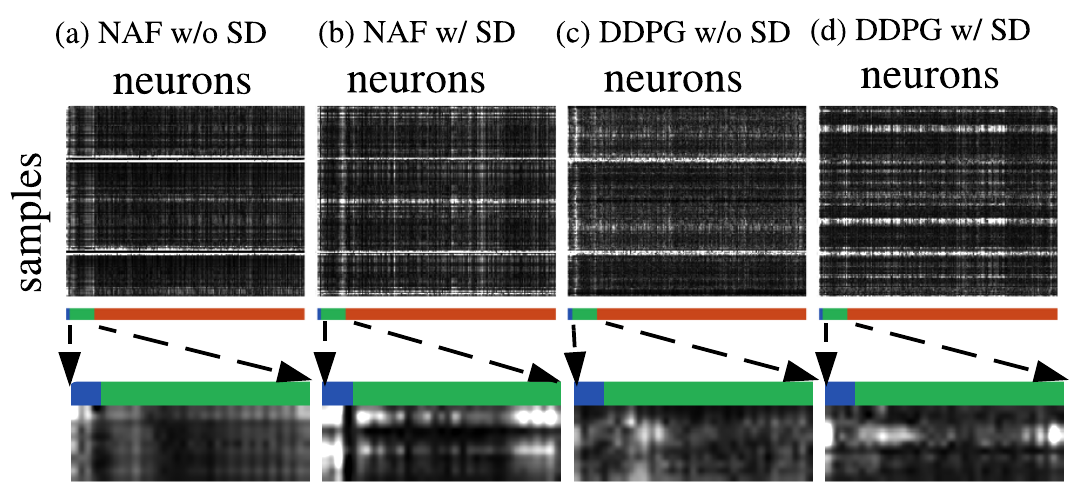}}
\subfigure[]{\label{fig:grad_exp_img}\includegraphics[width=0.48\columnwidth]{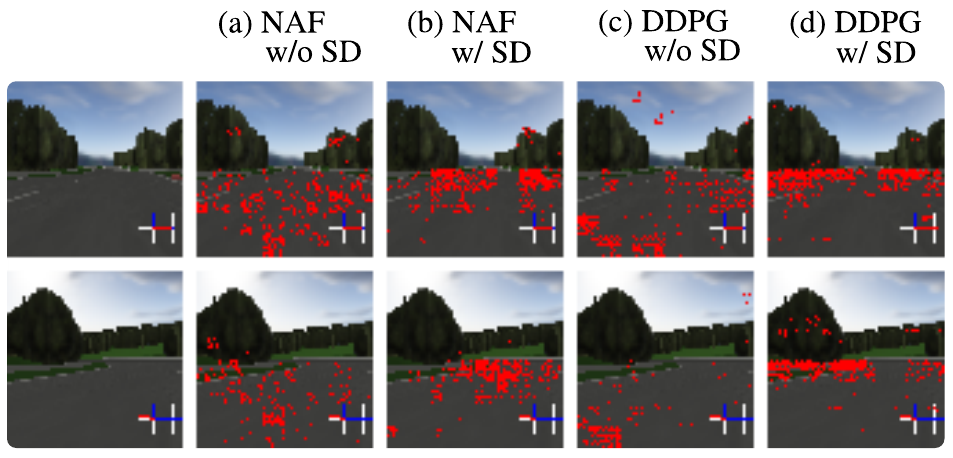}}
\caption{(a)The visualization of the magnitude of gradient for each neuron. The whiter color means the higher gradient. The color bar represents three different sensor modules: physical state(blue), Laser(green), and Image(red). (b) The gradient responses of actions on the image input for each of the multi-modal agents. The top $20\%$ gradients are marked red.}
\label{fig:grad_exp_all}
\vskip -0.15in
\end{figure}

\textbf{Visualize Policy Attention Region: }
The average gradient in the policy sensitivity section can also be used to visualize the regions among each sensor where the policy network pays attentions. As shown in Fig. \ref{fig:grad_exp}, we observe that policies trained with SD have higher gradients on neurons corresponding to the corner inputs of the laser sensor, indicating that a more sparse and meaningful policy is learned. These corner inputs corresponded to the laser beams that are oriented perpendicularly to the vehicle's direction of motion, and give an estimate of its relative position on the track. To look for similar patterns in Fig. \ref{fig:grad_exp_img}, image pixels with higher gradients are marked to interpret the policy's view of the world. We pick two scenarios, 1) straight track and 2) sharp left turn, depicted by the first and second rows in the figure. Note that though policies trained without SD tend to focus more on the road, those areas are in plain color and offer little salient information. In conclusion, policies trained with SD are more sensitive to features such as road boundary, which is crucial for long horizon planning. In comparison, networks trained without SD have relatively low and unclear gradients over both laser and image sensor state space.

\section{Conclusions and Future Work}
In this work, we introduce a new stochastic regularization technique called Sensor Dropout to promote an effective fusing of information from multiple sensors. The variance of the resulting policy can be further reduced by introducing an auxiliary loss during training. We show that SD reduces the policy sensitivity to a particular sensor subset, and guarantees functionality even in the face of a sensor subset failure. 
Moreover, the policy network is able to automatically infer and weight locations providing salient information. 
For future work, we wish to extend the framework to 
other environments such as real robotics systems, and other algorithms like TRPO \cite{TRPO}, and Q-Prop \cite{DBLP:journals/corr/GuLGTL16}, etc.. 
Secondly, systematic investigation into the problems such as how to augment the reward function for other important driving tasks like collision avoidance, and lane changing, and how to adaptively adjust the SD distribution during training are also interesting avenues that merit further study.

\section*{Acknowledgement}
The authors would like to thank Po-Wei Chou, Humphrey Hu,
and Ming Hsiao for many helpful discussions, suggestions and comments on the paper. This research was funded under award by Yamaha Motor Corporation.

\bibliographystyle{unsrtnat}
\bibliography{ref}

\begin{thebibliography}{34}
\providecommand{\natexlab}[1]{#1}
\providecommand{\url}[1]{\texttt{#1}}
\expandafter\ifx\csname urlstyle\endcsname\relax
  \providecommand{\doi}[1]{doi: #1}\else
  \providecommand{\doi}{doi: \begingroup \urlstyle{rm}\Url}\fi

\bibitem[Bojarski et~al.(2016)Bojarski, Del~Testa, Dworakowski, Firner, Flepp,
  Goyal, Jackel, Monfort, Muller, Zhang, et~al.]{nvidiacar}
M.~Bojarski, D.~Del~Testa, D.~Dworakowski, B.~Firner, B.~Flepp, P.~Goyal, L.~D.
  Jackel, M.~Monfort, U.~Muller, J.~Zhang, et~al.
\newblock End to end learning for self-driving cars.
\newblock \emph{arXiv preprint arXiv:1604.07316}, 2016.

\bibitem[Ross et~al.(2011)Ross, Gordon, and Bagnell]{ross2011reduction}
S.~Ross, G.~J. Gordon, and D.~Bagnell.
\newblock A reduction of imitation learning and structured prediction to
  no-regret online learning.
\newblock In \emph{AISTATS}, volume~1, page~6, 2011.

\bibitem[Mnih et~al.(2013)Mnih, Kavukcuoglu, Silver, Graves, Antonoglou,
  Wierstra, and Riedmiller]{mnih2013playing}
V.~Mnih, K.~Kavukcuoglu, D.~Silver, A.~Graves, I.~Antonoglou, D.~Wierstra, and
  M.~Riedmiller.
\newblock Playing atari with deep reinforcement learning.
\newblock In \emph{NIPS'13 Workshop on Deep Learning}, 2013.

\bibitem[Mnih et~al.(2015)Mnih, Kavukcuoglu, Silver, Rusu, Veness, Bellemare,
  Graves, Riedmiller, Fidjeland, Ostrovski, et~al.]{mnih2015human}
V.~Mnih, K.~Kavukcuoglu, D.~Silver, A.~A. Rusu, J.~Veness, M.~G. Bellemare,
  A.~Graves, M.~Riedmiller, A.~K. Fidjeland, G.~Ostrovski, et~al.
\newblock Human-level control through deep reinforcement learning.
\newblock \emph{Nature}, 518\penalty0 (7540):\penalty0 529--533, 2015.

\bibitem[Wymann et~al.(2000)Wymann, Espi{\'e}, Guionneau, Dimitrakakis, Coulom,
  and Sumner]{wymann2000torcs}
B.~Wymann, E.~Espi{\'e}, C.~Guionneau, C.~Dimitrakakis, R.~Coulom, and
  A.~Sumner.
\newblock Torcs, the open racing car simulator.
\newblock \emph{Software available at http://torcs. sourceforge. net}, 2000.

\bibitem[Gu et~al.(2016)Gu, Lillicrap, Sutskever, and Levine]{CDQN}
S.~Gu, T.~Lillicrap, I.~Sutskever, and S.~Levine.
\newblock Continuous deep q-learning with model-based acceleration.
\newblock In \emph{Proceedings of The 33rd International Conference on Machine
  Learning}, pages 2829--2838, 2016.

\bibitem[Lillicrap et~al.(2016)Lillicrap, Hunt, Pritzel, Heess, Erez, Tassa,
  Silver, and Wierstra]{DBLP:journals/corr/LillicrapHPHETS15}
T.~P. Lillicrap, J.~J. Hunt, A.~Pritzel, N.~Heess, T.~Erez, Y.~Tassa,
  D.~Silver, and D.~Wierstra.
\newblock Continuous control with deep reinforcement learning.
\newblock In \emph{International Conference on Learning Representations
  (ICLR)}, 2016.

\bibitem[Patel et~al.(2017)Patel, Choromanska, Krishnamurthy, and
  Khorrami]{patelsensor}
N.~Patel, A.~Choromanska, P.~Krishnamurthy, and F.~Khorrami.
\newblock Sensor modality fusion with cnns for ugv autonomous driving in indoor
  environments.
\newblock In \emph{International Conference on Intelligent Robots and Systems
  (IROS)}. IEEE, 2017.

\bibitem[Bohez et~al.(2017)Bohez, Verbelen, De~Coninck, Vankeirsbilck, Simoens,
  and Dhoedt]{bohez2017sensor}
S.~Bohez, T.~Verbelen, E.~De~Coninck, B.~Vankeirsbilck, P.~Simoens, and
  B.~Dhoedt.
\newblock Sensor fusion for robot control through deep reinforcement learning.
\newblock \emph{preprint arXiv:1703.04550}, 2017.

\bibitem[Urmson et~al.(2007)Urmson, Bagnell, Baker, Hebert, Kelly, Rajkumar,
  Rybski, Scherer, Simmons, Singh, et~al.]{multimodaltartan}
C.~Urmson, J.~A. Bagnell, C.~R. Baker, M.~Hebert, A.~Kelly, R.~Rajkumar, P.~E.
  Rybski, S.~Scherer, R.~Simmons, S.~Singh, et~al.
\newblock Tartan racing: A multi-modal approach to the darpa urban challenge.
\newblock 2007.

\bibitem[Cho et~al.(2014)Cho, Seo, Kumar, and Rajkumar]{cho2014multi}
H.~Cho, Y.-W. Seo, B.~V. Kumar, and R.~R. Rajkumar.
\newblock A multi-sensor fusion system for moving object detection and tracking
  in urban driving environments.
\newblock In \emph{International Conference on Robotics and Automation (ICRA)},
  pages 1836--1843. IEEE, 2014.

\bibitem[Qureshi et~al.(2016)Qureshi, Nakamura, Yoshikawa, and
  Ishiguro]{qureshi2016robot}
A.~H. Qureshi, Y.~Nakamura, Y.~Yoshikawa, and H.~Ishiguro.
\newblock Robot gains social intelligence through multimodal deep reinforcement
  learning.
\newblock In \emph{16th International Conference on Humanoid Robots}, pages
  745--751. IEEE, 2016.

\bibitem[Levine et~al.(2016)Levine, Finn, Darrell, and Abbeel]{levine2016end}
S.~Levine, C.~Finn, T.~Darrell, and P.~Abbeel.
\newblock End-to-end training of deep visuomotor policies.
\newblock \emph{Journal of Machine Learning Research}, 17\penalty0
  (39):\penalty0 1--40, 2016.

\bibitem[Mirowski et~al.(2017)Mirowski, Pascanu, Viola, Soyer, Ballard, Banino,
  Denil, Goroshin, Sifre, Kavukcuoglu, Kumaran, and Hadsell]{mirowski2017a}
P.~Mirowski, R.~Pascanu, F.~Viola, H.~Soyer, A.~Ballard, A.~Banino, M.~Denil,
  R.~Goroshin, L.~Sifre, K.~Kavukcuoglu, D.~Kumaran, and R.~Hadsell.
\newblock Learning to navigate in complex environments.
\newblock In \emph{International Conference on Learning Representations
  (ICLR)}, 2017.

\bibitem[Ngiam et~al.(2011)Ngiam, Khosla, Kim, Nam, Lee, and Ng]{ngmultimodal}
J.~Ngiam, A.~Khosla, M.~Kim, J.~Nam, H.~Lee, and A.~Y. Ng.
\newblock Multimodal deep learning.
\newblock In \emph{Proceedings of the 28th international conference on machine
  learning (ICML-11)}, pages 689--696, 2011.

\bibitem[Srivastava and Salakhutdinov(2012)]{srivastava2012multimodal}
N.~Srivastava and R.~R. Salakhutdinov.
\newblock Multimodal learning with deep boltzmann machines.
\newblock In \emph{Advances in neural information processing systems}, pages
  2222--2230, 2012.

\bibitem[Srivastava et~al.(2014)Srivastava, Hinton, Krizhevsky, Sutskever, and
  Salakhutdinov]{dropout}
N.~Srivastava, G.~E. Hinton, A.~Krizhevsky, I.~Sutskever, and R.~Salakhutdinov.
\newblock Dropout: a simple way to prevent neural networks from overfitting.
\newblock \emph{Journal of Machine Learning Research}, 15\penalty0
  (1):\penalty0 1929--1958, 2014.

\bibitem[Murdock et~al.(2016)Murdock, Li, Zhou, and Duerig]{blockout}
C.~Murdock, Z.~Li, H.~Zhou, and T.~Duerig.
\newblock Blockout: Dynamic model selection for hierarchical deep networks.
\newblock In \emph{Proceedings of the IEEE Conference on Computer Vision and
  Pattern Recognition}, pages 2583--2591, 2016.

\bibitem[Wan et~al.(2013)Wan, Zeiler, Zhang, Cun, and Fergus]{dropconnect}
L.~Wan, M.~Zeiler, S.~Zhang, Y.~L. Cun, and R.~Fergus.
\newblock Regularization of neural networks using dropconnect.
\newblock In \emph{Proceedings of the 30th International Conference on Machine
  Learning (ICML-13)}, pages 1058--1066, 2013.

\bibitem[Krueger et~al.(2016)Krueger, Maharaj, Kram{\'a}r, Pezeshki, Ballas,
  Ke, Goyal, Bengio, Larochelle, Courville, et~al.]{zoneout}
D.~Krueger, T.~Maharaj, J.~Kram{\'a}r, M.~Pezeshki, N.~Ballas, N.~R. Ke,
  A.~Goyal, Y.~Bengio, H.~Larochelle, A.~Courville, et~al.
\newblock Zoneout: Regularizing rnns by randomly preserving hidden activations.
\newblock \emph{arXiv preprint arXiv:1606.01305}, 2016.

\bibitem[Fraz{\~a}o and Alexandre(2014)]{dropall}
X.~Fraz{\~a}o and L.~A. Alexandre.
\newblock Dropall: Generalization of two convolutional neural network
  regularization methods.
\newblock In \emph{International Conference Image Analysis and Recognition},
  pages 282--289. Springer, 2014.

\bibitem[Neverova et~al.(2016)Neverova, Wolf, Taylor, and Nebout]{moddrop}
N.~Neverova, C.~Wolf, G.~Taylor, and F.~Nebout.
\newblock Moddrop: adaptive multi-modal gesture recognition.
\newblock \emph{IEEE Transactions on Pattern Analysis and Machine
  Intelligence}, 38\penalty0 (8):\penalty0 1692--1706, 2016.

\bibitem[Mnih et~al.(2016)Mnih, Badia, Mirza, Graves, Lillicrap, Harley,
  Silver, and Kavukcuoglu]{A3C}
V.~Mnih, A.~P. Badia, M.~Mirza, A.~Graves, T.~P. Lillicrap, T.~Harley,
  D.~Silver, and K.~Kavukcuoglu.
\newblock Asynchronous methods for deep reinforcement learning.
\newblock In \emph{International Conference on Machine Learning}, 2016.

\bibitem[Jaderberg et~al.(2016)Jaderberg, Mnih, Czarnecki, Schaul, Leibo,
  Silver, and Kavukcuoglu]{DBLP:journals/corr/JaderbergMCSLSK16}
M.~Jaderberg, V.~Mnih, W.~M. Czarnecki, T.~Schaul, J.~Z. Leibo, D.~Silver, and
  K.~Kavukcuoglu.
\newblock Reinforcement learning with unsupervised auxiliary tasks.
\newblock \emph{CoRR}, abs/1611.05397, 2016.
\newblock URL \url{http://arxiv.org/abs/1611.05397}.

\bibitem[Yoshida(2016)]{GymTORCS}
N.~Yoshida.
\newblock Gym-torcs.
\newblock \url{https://github.com/ugo-nama-kun/gym_torcs}, 2016.

\bibitem[Liaw et~al.()Liaw, Krishnan, Garg, Crankshaw, Gonzalez, and
  Goldberg]{meta_policy}
R.~Liaw, S.~Krishnan, A.~Garg, D.~Crankshaw, J.~E. Gonzalez, and K.~Goldberg.
\newblock Composing meta-policies for autonomous driving using hierarchical
  deep reinforcement learning.

\bibitem[Simonyan et~al.(2014)Simonyan, Vedaldi, and
  Zisserman]{simonyan2013deep}
K.~Simonyan, A.~Vedaldi, and A.~Zisserman.
\newblock Deep inside convolutional networks: Visualising image classification
  models and saliency maps.
\newblock In \emph{Proceedings of the IEEE Conference on Computer Vision and
  Pattern Recognition}, 2014.

\bibitem[Wang et~al.(2016)Wang, de~Freitas, and Lanctot]{WangFL15}
Z.~Wang, N.~de~Freitas, and M.~Lanctot.
\newblock Dueling network architectures for deep reinforcement learning.
\newblock In \emph{International Conference on Machine Learning (ICML)}, 2016.

\bibitem[Schulman et~al.(2015)Schulman, Levine, Abbeel, Jordan, and
  Moritz]{TRPO}
J.~Schulman, S.~Levine, P.~Abbeel, M.~I. Jordan, and P.~Moritz.
\newblock Trust region policy optimization.
\newblock In \emph{ICML}, pages 1889--1897, 2015.

\bibitem[Gu et~al.(2017)Gu, Lillicrap, Ghahramani, Turner, and
  Levine]{DBLP:journals/corr/GuLGTL16}
S.~Gu, T.~P. Lillicrap, Z.~Ghahramani, R.~E. Turner, and S.~Levine.
\newblock Q-prop: Sample-efficient policy gradient with an off-policy critic.
\newblock In \emph{International Conference on Learning Representations
  (ICLR)}, 2017.

\bibitem[Sutton et~al.(1999)Sutton, McAllester, Singh,
  et~al.]{sutton1999policy}
R.~S. Sutton, D.~A. McAllester, S.~P. Singh, et~al.
\newblock Policy gradient methods for reinforcement learning with function
  approximation.
\newblock In \emph{NIPS}, volume~99, pages 1057--1063, 1999.

\bibitem[Silver et~al.(2014)Silver, Lever, Heess, Degris, Wierstra, and
  Riedmiller]{dpg}
D.~Silver, G.~Lever, N.~Heess, T.~Degris, D.~Wierstra, and M.~Riedmiller.
\newblock Deterministic policy gradient algorithms.
\newblock In \emph{ICML}, 2014.

\bibitem[Lau(2016)]{BenLau16}
Y.-P. Lau.
\newblock Using keras and deep deterministic policy gradient to play torcs.
\newblock \url{https://yanpanlau.github.io/2016/10/11/Torcs-Keras.html}, 2016.

\bibitem[Kingma and Ba(2015)]{adam}
D.~P. Kingma and J.~L. Ba.
\newblock Adam: A method for stochastic optimization.
\newblock 2015.

\end{thebibliography}

\newpage

\textbf{{\huge Supplementary Material}}
\appendix

\section{Continuous Action Space Algorithms} 
\label{sec:supply_a}

\begin{figure}[b]
\begin{center}
\centerline{\includegraphics[width=0.6\columnwidth,trim= 80 900 110 70, clip=true]{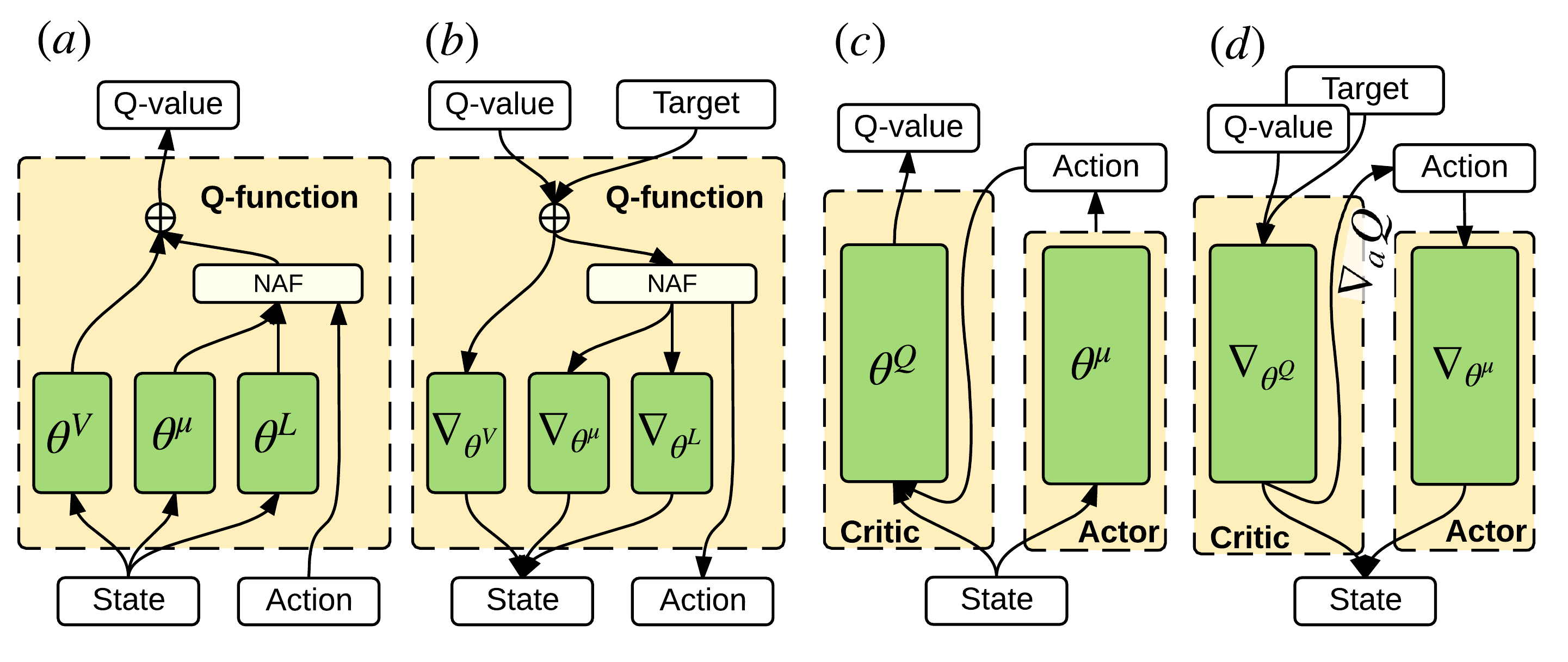}}
\caption{Schematic illustration of (a) forward and (b) back-propagation for NAF, and (c) forward and (d) back-propagation for DDPG. Green modules are functions approximated with Deep Nets.}
\label{fig:CDQN-DDPG}
\end{center}
\end{figure} 

\subsection{Normalized Advantage Function (NAF)} 
\label{sec:CDQN}

Q-learning \cite{sutton1999policy} is an off-policy model-free algorithm, where agent learns an approximated $Q$ function, and follows a greedy policy $\mu(s)=\arg\max_aQ(s,a)$ at each step. The objective function 
$J = \mathbb{E}_{s_i, r_i \sim E~, a_i \sim \pi}[R_1]$, can be reached by minimizing the square loss Bellman error $L = \frac{1}{N} \sum_i^N (y_i-Q(s_i,a_i|\theta^Q))^2$,
where target $y_i$ is defined as $r(s_i,a_i) + \gamma Q(s_{i+1},\mu(s_{i+1}))$.


Recently, \citep{CDQN} proposed a continuous variant of Deep Q-Learning by a clever network construction. The $Q$ network, which they called Normalized Advantage Function (NAF), parameterized the advantage function quadratically over the action space, and is weighted by non-linear feature of states. 
\begin{align}
\centering
Q(s,a|\theta^Q) &= A(s,a | \theta^\mu, \theta^L) + V(s|\theta^V) \\
A(s,a | \theta^\mu, \theta^L) &= -\frac{1}{2}(a-\mu(s|\theta^\mu))^T P(s|\theta^L)\nonumber \\
&\qquad \qquad \qquad(a-\mu(s|\theta^\mu)) \label{equ:NAF} \\
P(s|\theta^L) &= L(s|\theta^L)^TL(s|\theta^L) \label{equ:P}
\end{align}
During run-time, the greedy policy can be performed by simply taking the output of sub-network $a = \mu(s|\theta^\mu)$. The data flow at forward prediction and back-propagation steps are shown in Fig. \ref{fig:CDQN-DDPG} (a) and (b), respectively.

\subsection{Deep Deterministic Policy Gradient (DDPG)}
An alternative approach to continuous RL tasks was the use of an actor-critic framework, which maintains an explicit policy function, called \textit{actor}, and an action-value function called as \textit{critic}. 
In \cite{dpg}, a novel \emph{deterministic} policy gradient (DPG) approach was proposed and it was shown that deterministic policy gradients have a model-free form and follow the gradient of the action-value function. 
\begin{equation}
\nabla_{\theta^\mu} J = \mathbb{E}[\nabla_a Q(s,a|\theta^Q) \nabla_a \mu(s)]
\label{dpg}
\end{equation}
\citep{dpg} proved that using the policy gradient calculated in (\ref{dpg}) to update model parameters leads to the maximum expected reward.

Building on this result, \citep{DBLP:journals/corr/LillicrapHPHETS15} proposed an extension of DPG with deep architecture to generalize their prior success with discrete action spaces \cite{mnih2015human} onto continuous spaces. Using the DPG, an off-policy algorithm was developed to estimate the $Q$ function using a differentiable function approximator. Similar techniques as in \cite{mnih2015human} were utilized for stable learning. In order to explore the full state and action space, an exploration policy was constructed by adding Ornstein-Uhlenbeck noise process. 
The data flow for prediction and back-propagation steps are shown in Fig. \ref{fig:CDQN-DDPG} (c) and (d), respectively.

\section{Experiment Details}
\label{sec:supply_b}

\subsection{Exploration and Reward}
An exploration strategy is injected adding an Ornstein-Uhlenbeck process noise
to the output of the policy network. The choice of reward function is slightly different from  \citep{DBLP:journals/corr/LillicrapHPHETS15} and \citep{A3C} as an additional penalty term to penalize side-ways drifting along the track was added. In practice, this modification leads to more stable policies during training \cite{BenLau16}. 

\subsection{Network Architecture}

For laser feature extraction module, we use two $1D$ convolution layers with $4$ filters of size $4 \times 1$, while image feature extraction is composed of three $2D$ convolution layers: one layer of $16$ filters of size $4 \times 4$ and striding length $4$, followed by two layers each with $32$ filters of size $2 \times 2 $ and striding length $2$. Batch normalization is followed after every convolution layer. All these extraction modules are fused and are later followed up with two fully-connected layers of $200$ hidden units each. All hidden layers have \emph{relu} activations. The final layer of the critic network use \emph{leaner} activation, while the output of the actor network are bounded using \emph{tanh} activation. We use sigmoid activation for the output of $L$ network in NAF. In practice, it leads to a more stable training for high dimensional state space. We trained with minibatch size of $16$. 

We used Adam \cite{adam} for learning the network parameters. For DDPG, the learning rates for actor and critic are $10^{-4}$ and $10^{-3}$, respectively. We allow the actor and critic to maintain its own feature extraction module. In practice, sharing the same extraction module can lead to unstable training. Note that the NAF algorithm maintains three separate networks, which represent the value function ($V(s|\theta^V)$), policy network ($\mu(s|\theta^\mu)$), and the state-dependent covariance matrix in the action space ($P(s|\theta^L)$), respectively. In order to maintain a similar experiment setting and avoid unstable training, we maintain two independent feature extraction modules for $\theta^\mu$, and both $\theta^V$ and $\theta^L$. In a similar vein, we apply a learning rate of $10^{-4}$ for $\theta^\mu$, and $10^{-3}$ for both  $\theta^\mu$ and $\theta^V$. 

\begin{table}[t]
	\vskip 0.1in
	\caption{Model Specification}
	\label{table:model-spec}
	\vskip 0.1in
    \centering
    \begin{small}
    \begin{tabular}{ccc}
    \toprule 
    \centering
    Model ID & State Dimensionality & Description \\ \midrule \midrule
    Physical & 10 & \\
    Lasers & 4 $\times$ 19 & 4 consecutive laser scans \\
    Images & 12 $\times$ 64 $\times$ 64 & 4 consecutive RGB image \\
    Multi  & 10+1$\times$19+3$\times$64$\times$64 & all sensor streams at current time step \\ \toprule
    \end{tabular}
    \end{small}
\end{table}

\subsection{Simulated Sensor Detail}

\begin{figure}[t]
\centering
\vskip 0.2in
\includegraphics[width=0.7\columnwidth]{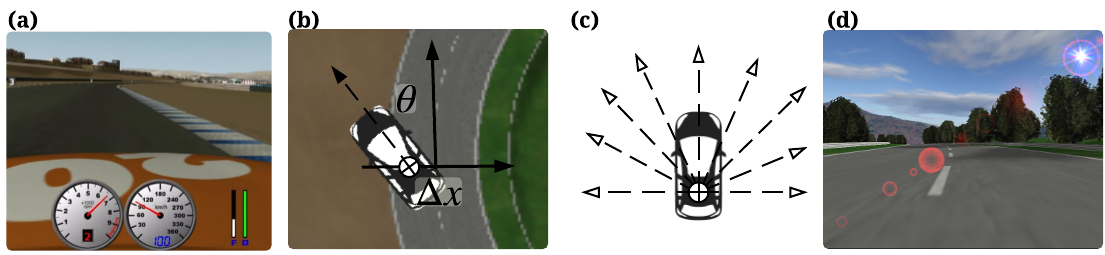} 
\caption{Sensors used in the TORCS racing car simulator: \textit{Sensor 1:} Physical information such as velocity (a), position, and orientation (b), \textit{Sensor 2:} Laser range finder (c), and \textit{Sensor 3:} Front-view camera (d). Sensor dimensionality details listed in Sec. \ref{sec:platform}.}
\label{fig:TORCS}
\end{figure} 

As shown in Fig. \ref{fig:TORCS}, the physical state is a $10$ DOF hybrid state, including $3$D velocity ($3$ DOF), position and orientation with respect to track center-line ($2$ DOF), and finally rotational speed of $4$ wheels ($4$ DOF) and engine ($1$ DOF). Each laser scan is composed of $19$ readings spanning a $180\degree$ field-of-view in the the front of car. Finally, camera provides RGB channels with resolution $64 \times 64$. 

\section{More Experimental Results}
\label{sec:supply_c}

\subsection{Effect of Auxiliary Loss}

\begin{table}[t]
	\vskip 0.1in
	\caption{Covariance of the first three Principal Component}
	\label{table:pca-3-covariance}
	\vskip 0.1in
    \centering
    \begin{small}
    \begin{sc}
    \begin{tabular}{c|ccc|ccc}
    \toprule 
    \centering
     & \multicolumn{3}{c}{NAF} & \multicolumn{3}{c}{DDPG}  \\
     Principal Component & w/oSD & w/SD & w/SD+aux & w/oSD & w/SD & w/SD+aux \\ \midrule \midrule
     First (\%) & 94.9  &  82.0  &  58.9 & 
                93.4  &  59.2  &  47.4 \\
     Second  (\%) & 4.1  &  12.3  &  25.2 & 
                    3.1  &  20.7  &  21.9 \\ 
	Third  (\%) & 0.6  &  3.1  &  5.3 & 
                1.6  &  6.2  &  6.1 \\ \toprule
    \end{tabular}
    \end{sc}
    \end{small}
\end{table}

\begin{table}[t]
	\vskip 0.1in
	\caption{Action Variation w.r.t. multimodal sensor}
	\label{table:pca-action-variance}
	\vskip 0.1in
    \centering
    \begin{small}
    \begin{sc}
    \begin{tabular}{c|ccc|ccc}
    \toprule 
    \centering
     & \multicolumn{3}{c}{NAF} & \multicolumn{3}{c}{DDPG}  \\
     & w/oSD & w/SD & w/SD+aux & w/oSD & w/SD & w/SD+aux \\ \midrule \midrule
     Steering & 0.1177  &  0.0819  &  \textbf{0.0135} & 
                0.3329  &  0.0302  &  \textbf{0.0290} \\
     Acceleration & 0.4559  &  0.0472  &  \textbf{0.0186} & 
                    0.5714  &  0.0427  &  \textbf{0.0143} \\ \toprule
    \end{tabular}
    \end{sc}
    \end{small}
\end{table}

The covariance of PCA and the actual action variance is summarized in Table \ref{table:pca-3-covariance} and \ref{table:pca-action-variance}, respectively.

\end{document}